\documentclass[letterpaper, 10pt, conference]{ieeeconf}
\pdfoutput=1
\IEEEoverridecommandlockouts 
\usepackage{import}
\usepackage{style}

\DeclareMathOperator*{\argmax}{argmax}
\newsavebox{\measurebox}
\usepackage{pifont}
\newcommand{\cmark}{\ding{51}}%
\newcommand{\xmark}{\ding{55}}%

\usepackage{xcolor}
\usepackage{colortbl}
\usepackage{footmisc}

\makeatletter
\let\NAT@parse\undefined
\makeatother
\usepackage[pdfborder={0 0 1}]{hyperref}  

\title{\LARGE \bf
    Reactive Stepping for Humanoid Robots using Reinforcement Learning: Application to Standing Push Recovery on the Exoskeleton Atalante
}

\author{Alexis Duburcq$^{1,2,3}$, Fabian Schramm$^{1}$, Guilhem Bo\'{e}ris$^{1}$, Nicolas Bredeche$^{2}$ and Yann Chevaleyre$^{3}$%
\thanks{$^{1}$Wandercraft, Paris, France. \href{mailto:alexis.duburcq@gmail.com}{\nolinkurl{<alexis.duburcq@gmail.com>}}
}%
\thanks{$^{2}$Sorbonne Universit\'{e}, CNRS, Institut des Syst\`{e}mes Intelligents et de Robotique, ISIR, F-75005 Paris, France.
}%
\thanks{\hbadness=2100 $^{3}$Universit\'{e} Paris-Dauphine, PSL, CNRS, Laboratoire d'analyse et modélisation de systèmes pour l'aide à la décision, Paris, France.
}}

\begin{document}
\bstctlcite{BSTcontrol}

\clubpenalty=9996
\widowpenalty=9999 
\brokenpenalty=4991
\predisplaypenalty=10000
\postdisplaypenalty=1549
\displaywidowpenalty=1602

\maketitle


\begin{abstract}

State-of-the-art reinforcement learning is now able to learn versatile locomotion, balancing and push-recovery capabilities for bipedal robots in simulation. Yet, the reality gap has mostly been overlooked and the simulated results hardly transfer to real hardware. Either it is unsuccessful in practice because the physics is over-simplified and hardware limitations are ignored, or regularity is not guaranteed, and unexpected hazardous motions can occur.
This paper presents a reinforcement learning framework capable of learning robust standing push recovery for bipedal robots that smoothly transfer to reality, providing only instantaneous proprioceptive observations. By combining original termination conditions and policy smoothness conditioning, we achieve stable learning, sim-to-real transfer and safety using a policy without memory nor explicit history. Reward engineering is then used to give insights into how to keep balance.
We demonstrate its performance in reality on the lower-limb medical exoskeleton Atalante.

\end{abstract}


\section{Introduction}

Achieving dynamic stability for bipedal robots is one of the most complex tasks in robotics. Continuous feedback control is required to keep balance since the vertical posture is inherently unstable~\cite{Caron18StairClimb}. However, hybrid high-dimensional dynamics, kinematic redundancy, model and environment uncertainties, and hardware limitations make it hard to design robust embedded controllers. Offline trajectory planning for bipedal robots has been solved successfully through whole-body optimization~\cite{MotionPlanning13,Kuindersma2016optim}. In particular, stable walking on flat ground and without disturbances was achieved on the exoskeleton Atalante~\cite{Gurriet18Wdc}. Yet, online re-planning, robustness to uncertainties and emergency recovery in case of external perturbations are still very challenging.

Modern control approaches require a lot of expert knowledge and effort in tuning because of discrepancies between approximate models and reality. Solutions are mainly task-specific, and improving versatility is usually done by stacking several estimation and control strategies in a modular hierarchical architecture~\cite{Moro2018, Herzog14, Kim20}. Though efficient in practice, it makes the analysis and tuning increasingly difficult and thereby limits its capability. In contrast, deep reinforcement learning (RL) methods require expert knowledge and extensive efforts to tailor the optimization problem, rather than structuring explicit controllers and defining good approximate models. RL aims at solving observation, planning and control as a unified problem by end-to-end training of a policy~\cite{sutton2018reinforcement}. Tackling the problem globally maximizes the potential of this method, but state-of-the-art algorithms still face learning instabilities and difficulties in discovering satisfactory behaviors for practical applications.

A ubiquitous problem of controllers trained with deep RL is the lack of safety and smoothness. This is dangerous for real-life deployment as human beings cannot anticipate future motions. Without special care, the control varies discontinuously like a bang-bang controller, which can result in a poor transfer to reality, high power consumption, loud noise and system failures\cite{Mysore21CAPS}. Despite those potential limitations, a robust gait and push recovery for the bipedal Cassie robot was recently learned in simulation using deep RL and transferred successfully to the real device\cite{Li21PushCassie}. Concurrently, several works on standing push recovery for humanoid robots trained in simulation suggest that the approach is promising \cite{Ferigo21HumanoidPush, Melo20Push}, although the same level of performance has not been achieved on real humanoid robots yet.

\begin{figure}[t]
    \centering
    \vspace{-0.4em}
    \subfloat[front push shoulder\label{fig:mini1}]{%
    \includegraphics[height=23.2mm]{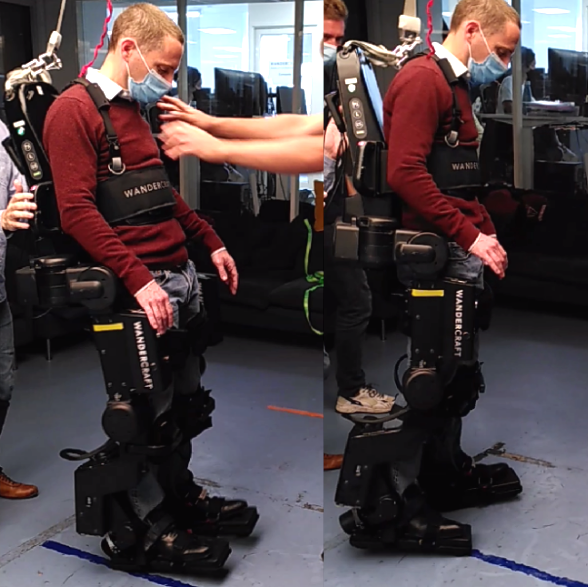}
    \hspace{0.25em}%
    \includegraphics[height=23.2mm]{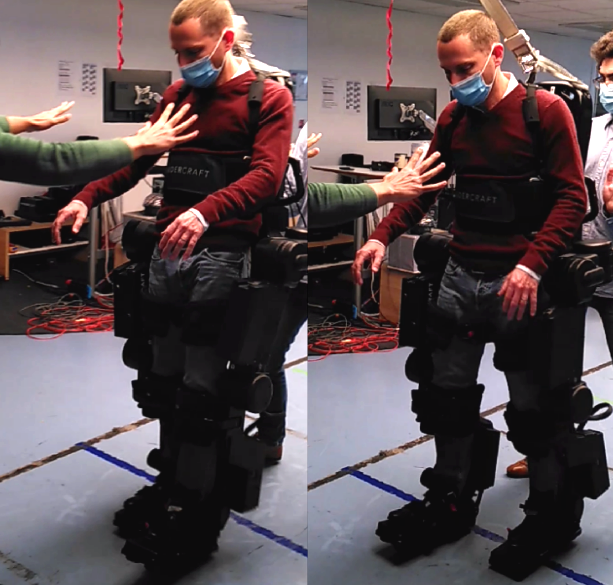}}
    \hfill
    \subfloat[strong kick pelvis\label{fig:mini3}]{%
    \includegraphics[trim=0 0 -0.2em 0, height=23.2mm]{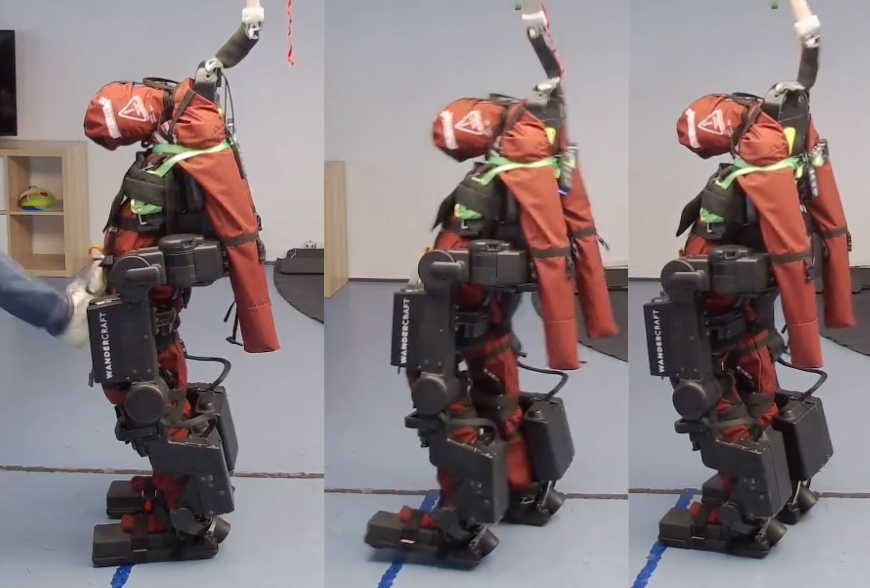}}
    \\[-0.6em]
    \subfloat[random user moves\label{fig:mini4}]{%
    \includegraphics[height=23.2mm]{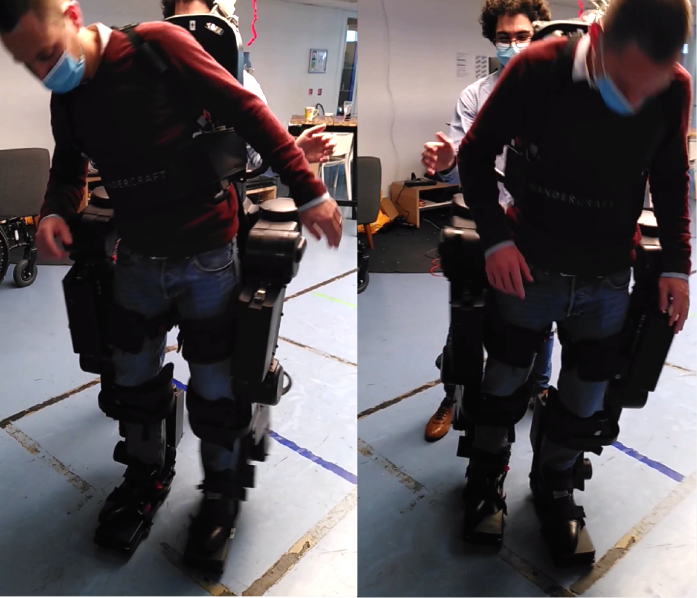}}
    \hfill
    \subfloat[side push hip\label{fig:mini5}]{%
    \includegraphics[trim=0 0 -0.1em 0, height=23.2mm]{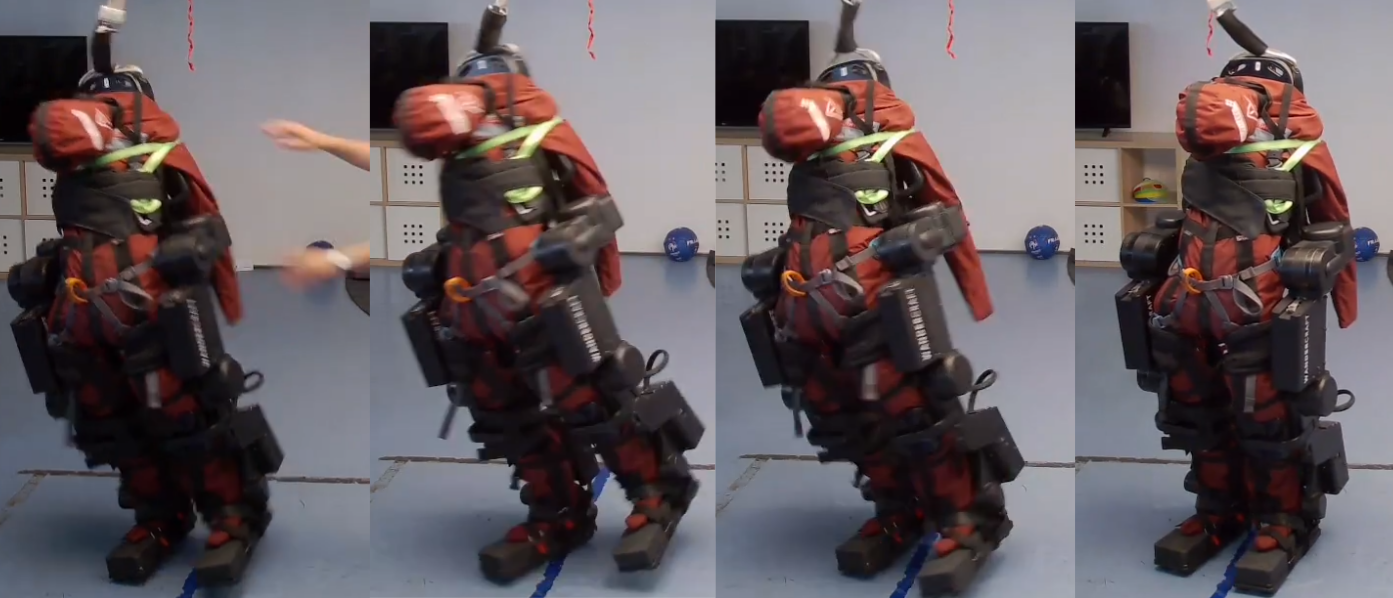}}
    \\[0.3em]
    \caption{Scenarios showing the robustness of the push recovery policy: smoothly transferred to different users, exoskeleton setups and force profiles.}
    \vspace{-1.2em}
    \label{fig:mini_scenarios}
\end{figure}

Our main contribution is the development of a purely reactive controller for standing push recovery on legged robots using deep RL, which is used as the last resort fallback in case of emergency. Precisely, we design an end-to-end policy featuring a variety of balancing strategies from the latest proprioceptive sensor data only, while promoting predictable, safe and smooth behavior. Our method combines carefully designed termination conditions, reward shaping and so-called policy smoothness conditioning \cite{Mysore21CAPS}:
\begin{itemize}
    \vspace{-0.2em}
    \setlength{\itemsep}{0pt}
    \setlength{\parskip}{0pt}
    \setlength{\parsep}{0pt}
    \item Termination conditions improve sample efficiency and learning stability by limiting the search space. Besides, they can enforce hard constraints such as hardware limitations and safety. We can generate sensible policies regardless of the reward function.
    \item Reward engineering improves exploration efficiency and promotes more natural behaviors.
    \item Smoothness conditioning favors few versatile strategies over many locally optimal behaviors. Slow motions are less sensitive to modeling errors including the internal dynamics of the feedback loop, so it alleviates the need for transfer learning approaches known for trading performance over robustness, e.g.\ domain randomization.
\end{itemize}
Some emerging strategies would be very challenging to reproduce using classical model-based control. Moreover, the policy can be directly transferred to a real robot. We demonstrate experimentally safe and efficient recovery behaviors for strong perturbations on the exoskeleton Atalante carrying different users, as shown in the supplementary video\footnote{\url{https://youtu.be/HLx6CHfpmBM}\label{fn:video}}.

\section{Related Work}

\subsection{Classical Non-Linear Control}

Upright standing offers a large variety of recovery strategies that can be leveraged in case of emergency to avoid falling down, among them: ankle, hip, stepping, height modulation and foot-tilting for any legged robot, plus angular momentum modulation for humanoid robots \cite{Yuan20BalanceControl}. For small perturbations, in-place recovery strategies controlling the Center of Pressure (CoP) \cite{Hyon09}, the centroidal angular momentum \cite{Stephens10}, or using foot-tilting\cite{Li17FootTilt, Caron19} are sufficient. To handle stronger perturbations, controllers based on Zero-Moment Point (ZMP) trajectory generation have been proposed \cite{Kajita03}, along with Model Predictive Control (MPC) methods controlling the ZMP \cite{MPC06Wieber}, but in practice the efficiency was limited. More recently, approaches based on the Capture Point (CP) showed promising results on real robots \cite{Caron18StairClimb}. The Linear Inverted Pendulum (LIP) model \cite{3DLip} is used to make online computations tractable. However, due to partial model validity, it tends to be restricted to moderately fast and conservative motions.

\subsection{Deep Reinforcement Learning}

Several ground-breaking advances were made in deep RL during the last decade. It has shown impressive effectiveness at solving complex continuous control tasks for toy models~\cite{LillicrapHPHETS15, HeessTSLMWTEWER17}, but real-world applications are still rare. Lately, deep RL was used to learn locomotion policies for dynamic and agile maneuvers on the quadruped ANYmal, which were smoothly transferred to reality \cite{Hwangbo2019, Miki2022LearningRP}. Besides, extremely natural motions were obtained on various simplified models by leveraging reference motion data from motion capture in an imitation learning framework \cite{Peng18DeepMimic,Peng2020}. Walking on flat ground, learned without providing any reference trajectories, was demonstrated on a mid-size humanoid \cite{Rodriguez2021}. However, the motion was slow and unnatural, with limited robustness to external forces. Promising results were achieved in simulation by several authors concurrently regarding standing push recovery \cite{Ferigo21HumanoidPush, ValkryStandingPush20, Yang20HumanBias}.
Yet, robust locomotion and standing push recovery for humanoid robots using deep RL falls short from expectations on real devices. The emerging behaviors are often unrealistic or hardly transfer to reality.

\subsection{Simulation to Real World Transfer}

Various policies have been trained in simulation for the bipedal robot Cassie and successfully transferred to reality. In~\cite{Xie2019}, stable walking has been achieved without disturbances based on reference trajectories generated using a traditional model-based method. Domain randomization is not needed because the policy predicts targets for a low-level PD controller and the simulation is faithful. The latter cannot be expected for an exoskeleton because of the unknown user dynamics. Concurrently, it was generalized to a whole gait library~\cite{Li21PushCassie} and no reference at all~\cite{CassieStairClimb}. Domain randomization enables to deal with disturbances for which it was never trained, including pushes. Although effective, it leads to more conservative behaviors than necessary. It can even prevent learning anything if the variability is not increased progressively~\cite{Li21PushCassie}. Alternatively, the stability can be improved by predicting high-level features for a model-based controller~\cite{Castillo2021RobustFM}, but it bounds the overall performance. Besides, a memory network or a history of previous timesteps is often used to thwart partial observability of the state~\cite{Li21PushCassie,CassieStairClimb}. It improves robustness to noise and model uncertainty, but it makes the training significantly more difficult~\cite{Miki2022LearningRP}.

The efficiency of RL-based controllers is limited in practice by the lack of smoothness in the predicted actions, which can be mitigated by filtering. Although it can be sufficient \cite{Li21PushCassie}, it is known to make policies unstable or underperforming \cite{Mysore21CAPS}. In contrast, the Conditioning for Action Policy Smoothness (CAPS) method adjusts the smoothness of the policy itself \cite{Mysore21CAPS}, by combining temporal and spatial regularization of the learning problem. It enables transfer to reality without domain randomization. Nevertheless, the formulation \cite{Mysore21CAPS} has flaws that we address in the current work: First, it penalizes exploration, impeding the learning algorithm to converge to the optimal solution. Second, it prevents bursts of acceleration which are necessary to recover from strong pushes, leading to an underperforming policy.

\begin{figure}[t]
    \centering\hspace*{\fill}
    \subfloat[Jiminy simulator\label{fig:jiminy}]{%
    \includegraphics[trim=-1.0em 0 -1.0em 2.0em, height=45.0mm]{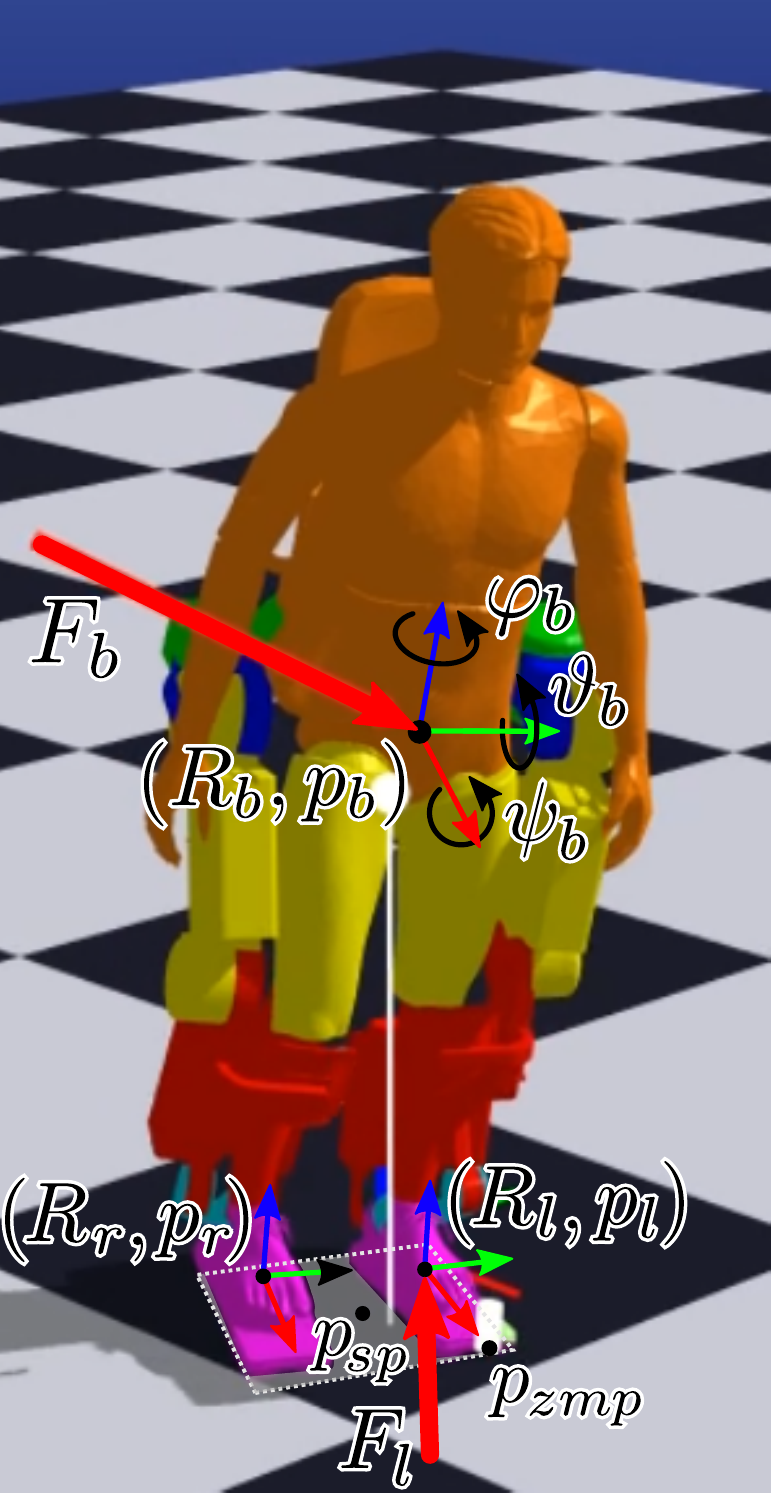}}
    \hfill
    \subfloat[Actuators and sensors\label{fig:hardware}]{%
    \includegraphics[trim=0 0.2em 0 3.3em, height=45.0mm]{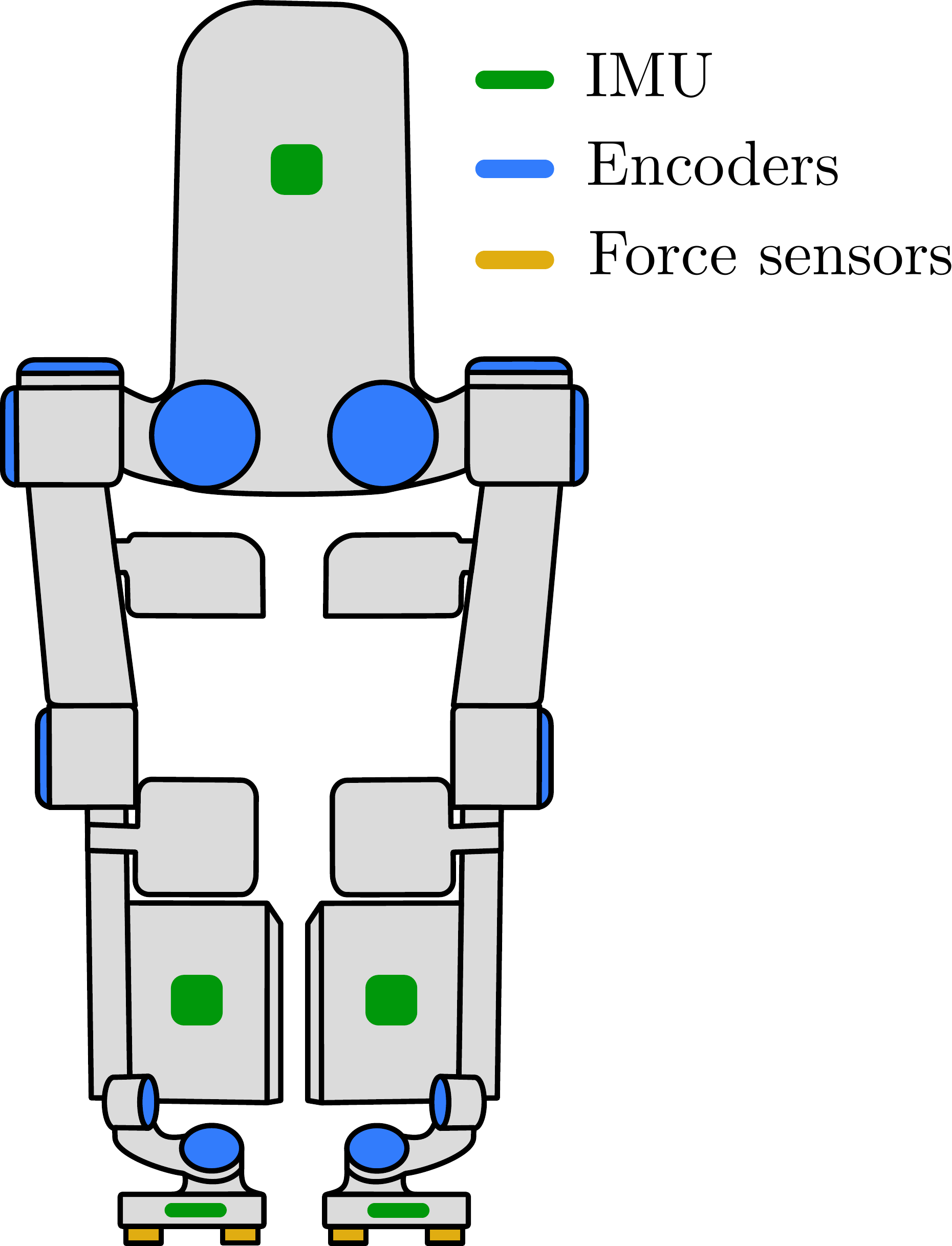}}\hspace*{\fill}
    \vspace{0.3em}
    \caption{Atalante in learning environment and hardware overview.}
    \vspace{-1.2em}
    \label{fig:atalante}
\end{figure}

\section{Background}

\subsection{The Reinforcement Learning Problem}

RL methods aim at training an agent to solve an episodic task through an iterative process of trial and error. The agent interacts with its environment in a discrete-time setting: at each time step $t$, the agent observes $o_t$ based on the state of the system $s_t$, receives a reward $r_t$ and subsequently performs an action $a_t$ given this information. It then moves to the next state $s_{t+1}$ and receives the reward $r_{t+1}$ associated with the transition $(s_t, a_t, s_{t+1})$. The interaction generates trajectories $\tau = \{(s_i,a_i,r_i)\}_{i=0}^T$ whose length $T$ depends on termination conditions. Formally, this problem can be described as an infinite-horizon Markov Decision Process (MDP)\cite{sutton2018reinforcement}. It is defined by a state space $S \in \mathbb{R}^n$, observation space $O \in \mathbb{R}^l$, valid action space $A \in \mathbb{R}^m$, transition function $p : S \times A \times S \rightarrow [0, 1]$, observation function $o: S \times A \rightarrow O$, reward function $r: S \times A \times S \rightarrow \mathbb{R}$ and discount factor $\gamma \in [0,1]$. The transition function $p$ encodes the dynamics of the agent in the world.

The agent's behavior is determined by a stochastic policy $\pi(\cdot | s_t)$ mapping states to distributions over actions $Pr(A)$. The goal is to find a policy $\pi^*$ maximizing the return $R(\tau)=\sum_{t=0}^{\infty} \gamma^{\;t} r_t$, i.e. the discounted cumulative reward, in expectation over the trajectories $\tau$ induced by the policy:
\begin{equation*}
    \pi^* = \argmax_{\pi} J(\pi) \text{ s.t. } J(\pi) = \mathbb{E}_{\tau \sim \pi} \left[R(\tau)\right] .
\end{equation*}

The discount factor enables a trade-off between long-term vs. short-term profit, which is critical for locomotion tasks.

\subsection{Policy Gradient Methods}

In deep RL, the policy is a neural network with parameters $\theta$. A direct gradient-based optimization of $\theta$ is performed over the objective function $J(\pi_{\theta})$. The analytical gradient of the return is unknown, so it computes an estimate of $\nabla_{\theta} J(\pi_{\theta})$ and applies a stochastic gradient ascent algorithm.
It yields
\begin{equation*}
    \nabla_{\theta} J (\pi_{\theta}) = \hat{\mathbb{E}}_{\tau \sim \pi_{\theta}} \left[\nabla_{\theta} \log(\pi_{\theta}(a_t | s_t)) A_t\right],
\end{equation*}
where $\hat{\mathbb{E}}$ is the expectation over a finite batch of trajectories and $A_t = R_t - V(s_t)$ is the advantage function corresponding to the difference between the future return $R_t = \sum_{k=0}^{\infty} \gamma^{\;k} r_{t+k}$ and the value function $V(s_t) = \hat{\mathbb{E}}_{\tau \sim \pi} [R_t | s_t]$.

Usually, the value function $V(s_t)$ is a neural network itself, trained in conjunction with the policy. Using it as a baseline reduces the variance of the gradient estimator. However, the gradient estimator still suffers from high variance in practice, weakening the convergence and asymptotic performance.

Up to now, Proximal Policy Optimization (PPO)~\cite{PPOSchulmanWDRK17} is the most reliable policy gradient algorithm for continuous control tasks. It is an actor-critic algorithm, which means that both the policy and the value function are trained. The policy is normally distributed with parametric mean $\mu_\theta$ but fixed standard deviation $\sigma$. It tackles learning instability by introducing a surrogate that gives a conservative gradient estimate bounding how much the policy is allowed to change:
\begin{equation*}
    L^{CLIP}(\theta) = \hat{\mathbb{E}}_t \left[\min \left(r_t(\theta) \hat{A}_t, \text{clip}\left(r_t(\theta), 1-\epsilon, 1+\epsilon\right)\hat{A}_t \right) \right]
    \label{eq:PPO}
\end{equation*}
where $r_t(\theta) = \frac{\pi_{\theta}(a_t | s_t)}{\pi_{\theta_{old}}(a_t|s_t)}$ is the likelihood ratio and $\hat{A}_t$ is the Generalized Advantage Estimator (GAE) \cite{schulman2018highdimensional}. The latter offers an additional parameter $\lambda$ over the naive estimate $A_t$ to adjust the bias vs. variance trade-off.

\section{Methodology}

\subsection{Learning Environment}

\subsubsection{Atalante Model}

Atalante is a crutch-less lower-limb exoskeleton for people with disabilities. It has 6 actuators on each leg and a set of basic proprioceptive sensors, see \autoref{fig:hardware}. The patient is rigidly fastened to the exoskeleton, which features dimensional adjustments to fit the individual morphology. In this regard, the system exoskeleton-patient can be viewed as a humanoid robot after merging their respective mass distributions. Contrary to usual humanoids, the control of the upper body is partial. For learning, a single average patient model is used.

We use the open-source simulator Jiminy \cite{Jiminy} based on Pinocchio \cite{pinocchioweb}, see \autoref{fig:jiminy}. It was originally created for classic control robotic problems and realistic simulation of legged robots. Jiminy includes motor inertia and friction, sensor noise and delay, constraint-based contact forces \cite{MuJoCo}, as well as an advanced mechanical flexibility model \cite{Vigne20}.

\begin{figure}
    \centering
    \includegraphics[trim=0 0 0 -1.0em, width=\columnwidth]{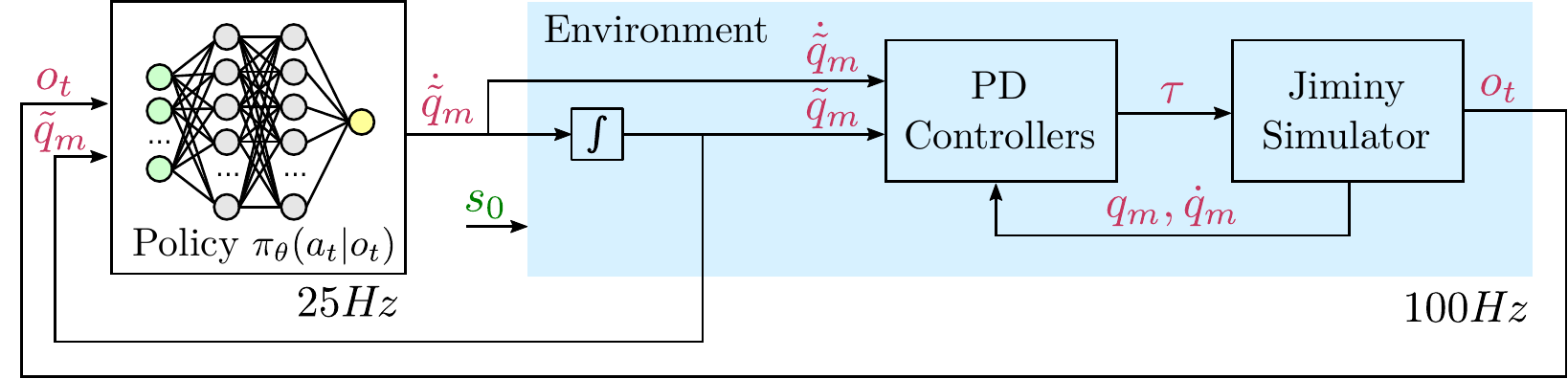}
    \caption{Overview of the proposed control system.}
    \vspace{-1.2em}
    \label{fig:control_system}
\end{figure}

\subsubsection{Action Space}

A distinctive feature in RL is the slow update frequency in contrast to classic control approaches, giving enough time for the effect of the actions to build up. First, it is critical for exploration efficiency, which relies on random action sampling. Second, it reduces the variance of the gradient estimate by increasing the signal-to-noise ratio.

The policy is trained to predict targets that are forwarded to decentralized low-level PD controllers running at higher frequency. Decoupled control of each joint is well-known to be robust to model uncertainties. Thus, this hybrid control architecture improves robustness and transferability compared to predicting motor torques $u$ directly. Moreover, these controllers can be tuned to trade off tracking accuracy versus compliance, smoothing out vibrations and errors in the predicted actions.
In this work, the policy predicts target motor velocities at 25 Hz, which are further integrated to target motor positions to enforce a smooth behavior.
We forward consistent target motor positions and velocities to the low-level PD controllers running at 100 Hz, see \autoref{fig:control_system}. It is necessary for accurate tracking, especially for dynamic motions. This consistency issue was always disregarded in related works since it does not affect performance but rather predictability and safety.

\subsubsection{State Space and Observation Space}

The user is not modeled explicitly but rather considered as an external disturbance. Hence, the theoretical state of the system $s_t$ is fully determined by the position $p_b$, roll-pitch-yaw orientation $\psi_b, \theta_b, \phi_b$, linear velocity $v_b$ and angular velocity $\omega_b$ of the pelvis of the robot, plus the motor positions $q_m$ and velocities $\dot{q}_m$. Even so, the state $s_t$ is only partially observable. The observation $o_t \in \mathcal{O}$ $\in \mathbb{R}^{49}$ is computed from low-cost proprioceptive sensors: one inertial measurement unit (IMU) estimating the roll, pitch and angular velocity of the pelvis, 8 vertical force sensors in both feet $F^z_{r,l}$ and 12 motor encoders measuring the motor positions. The motor velocities $\dot{q}_m$ are obtained by numerical differentiation. Additionally, the target motor positions $\tilde{q}_m^{t-1}$ from the last time step are included in the observation. No window over several time steps is aggregated. The quantities are independently normalized over training batches. Their distributions change over training iterations since it depends on the current policy, hence it would be hazardous to do it manually.

Some insightful quantities cannot be reliably estimated without exteroceptive sensors, e.g. the pelvis height $z_b$ and linear velocity $v_b$. They are not included in the observation space because any significant mismatch between simulated and real data may prevent transfer to reality. Although the odometry pose $p_o = [x_b, y_b, \varphi_b]^T$ is not observable, it is not blocking as the recovery strategies should be invariant to it.

\subsubsection{External Disturbances Modeling}

To learn sophisticated recovery strategies, the external pushes in the learning environment need to be thoughtfully scheduled. They must be strong enough to sometimes require stepping, but pushing too hard would prohibit learning. As suggested in \cite{Ferigo21HumanoidPush}, we apply forces of constant magnitude for a short duration periodically on the pelvis, where the orientation is sampled from a spherical distribution. In this work, the pushes are applied every 3s, with a jitter of 2s to not overfit to a fixed push scheme and learn recovering consecutive pushes. The pushes are bell-shaped instead of uniform for numerical stability. They have a peak magnitude of $\max \|F_b\|_2 = 800$N and are applied during 400ms. Premature convergence to a suboptimal policy was observed if the force is gradually increased from 0 to 800N over the episode duration instead.

\subsection{Initialization}
\label{subsec:init}

The initial state distribution $\rho_0(s_0) : S \rightarrow [0,1]$ defines the probability of an agent to start the episode in state $s_0$. It must ensure the agent has to cope with a large variety of situations to promote robust recovery strategies. Ideally, it should span all recoverable states, but this set is unknown as it depends on the optimal policy. Naive random sampling would generate many unrecoverable states or even trigger termination conditions instantly. Such trajectories have limited value and can make learning unstable.

We propose to sample the initial state uniformly among a finite set of reference trajectories $t \rightarrow (\hat{q}(t), \hat{\dot{q}}(t))$, and then to add Gaussian noise to increase the diversity. The trajectories correspond to the range of motions in everyday use: standing, walking forward or backward at different speeds and turning. They are generated beforehand using a traditional model-based planning framework for the average patient morphology~\cite{Gurriet18Wdc}. Therefore, the corresponding states are feasible and close to dynamic stability.

\subsection{Termination Conditions}

Bounding quantities in simulation is not helping to enforce hard constraints in reality unlike termination conditions. The latter stop the reward from accumulating. Assuming the reward must be always positive, it caps the return to a fraction of its maximum. The agent becomes risk-averse: being confident about preventing critical failure in a stochastic world requires extra caution. Otherwise, it is detrimental because the agent tends to kill itself on purpose. Terminal conditions are the counterpart to log barrier penalties in constrained optimization. It is complementary to strict safety guarantees at runtime and does not allow for getting rid of them.

One key contribution of this article is a set of carefully designed termination conditions. First, they ensure transfer to reality and safety. Secondly, they reduce the search space to sensible solutions only. Some local minima of poor recovery strategies are strongly discouraged from the start, which leads to more stable learning and faster convergence~\cite{ma2021spacetimeBounds}. Numerical values are robot-specific. Unless stated otherwise, they were obtained by qualitative study in simulation.

\subsubsection{Pelvis Height and Orientation}

We restrict the pose of the pelvis to avoid frightening the user and people around,
\begin{equation*}
    z_b > 0.3, \mkern12mu -0.4 < \psi_b < 0.4, \mkern12mu -0.25 < \vartheta_b < 0.7\;.
\end{equation*}

\subsubsection{Foot Collisions}

For safety, foot collision needs to be forestalled as it can hurt the patient and damage the robot,
\begin{equation*}
    \mathcal{D}(CH_{r}, CH_{l}) > 0.02,
\end{equation*}
where $CH_r,CH_l$ are the convex hulls of the right and left footprints respectively, and $\mathcal{D}$ is the euclidean distance.

\subsubsection{Joint Bound Collisions}

Hitting the mechanical stops $q^-,q^+$ is inconvenient but forbidding it completely is not desirable. Considering PD controllers and bounded torques, it induces safety margins that constrain the problem too strictly. It would impede performance while avoiding falling is the highest priority. Still, the impact velocity must be restricted to prevent destructive damage or injuries. An acceptable threshold has been estimated from real experimental data,
\begin{equation*}
    |\dot{q}_i| < 0.6 \text{ or } q^-_i < q_i < q^+_i.
\end{equation*}

\subsubsection{Reference Dynamics}

Long-term drift of the odometry position is inevitable, but it must be limited. The user is only supposed to rectify it occasionally and the operational space is likely to be confined in practice. We restrict the odometry displacement over a time window $\Delta p_o$ instead of its instantaneous velocity~\cite{ma2021spacetimeBounds}. It limits the drift while allowing a few recovery steps. In general, the expected odometry displacement must be subtracted if non-zero.
\begin{equation*}
    |\Delta p_o - \Delta \hat{p}_o| < [2.0, 3.0, \pi/2],
\end{equation*}
where $\Delta \star = \star(t) - \star(t-\Delta T) $ and $\Delta T = 20$s.

\subsubsection{Transient Dynamics}

The robot must track the reference if there is no hazard, only applying minor corrections to keep balance. Rewarding the agent for doing so is not effective as favoring robustness remains more profitable. Indeed, it would anticipate disturbances, lowering its current reward to maximize the future return, primarily averting termination. We need to allow large deviations to handle strong pushes but also urge to quickly cancel it afterwards.
\begin{equation*}
    \min_{t' \in [t-\Delta T, t]} \left\|q_m(t') - \hat{q}_m(t') \right\|_2 < 0.3 \; , \; \Delta T = 4\text{s}.
\end{equation*}

\subsubsection{Power Consumption}
We limit the power consumption to fit the hardware capability and avoid motor overheating,
\begin{equation*}
    P = \langle u, \dot{q}_m \rangle \; < 3\text{kW}.
\end{equation*}

\subsection{Reward Engineering}

We designed a set of cost components that provides insight into how to keep balance. Additionally, we use them as a means to trigger reactive steps only if needed, as it is a last resort emergency strategy. We aim to be generic, so they can be used in conjunction with any reference trajectory to enhance stability and provide recovery capability. The total reward is a normalized weighted sum of the individual costs
\begin{equation*}
    r = \sum\nolimits_i w_i K(c_i),
\end{equation*}
where $c_i$ is a cost component, $w_i$ its weight and $K$ a kernel function that scales them equally. We use the Radial Basis Function (RBF) with cutoff parameters $\kappa_i$
\begin{equation*}
    K(c_i) = \exp(-\kappa_i c_i^2) \; \in [0, 1].
\end{equation*}
The gradient vanishes for both very small and large values as a side effect of this scaling. The cutoff parameter $\kappa$ is used to adjust the operating range of every reward component.

\subsubsection{Reference Dynamics}

We encourage tracking a specific reference trajectory, which boils down to a stable resting pose for standing push recovery. Evaluating the tracking discrepancy by computing the absolute motor position error wrt a reference has several limitations:
\begin{itemize}
    \item Tendency to diverge because of high-dimensionality. One large scalar error is enough for the whole gradient to vanish because of the kernel.
    \item Inability to enforce high-level features independently, e.g. the position and orientation of the feet.
    \item Quantities like $p_o$ are allowed to drift, which requires using velocity error instead of position.
\end{itemize}
To overcome them, we extract a set of independent high-level features and define reward components for each of them. As they are unrelated, even if one feature is poorly optimized, it is possible to improve the others.

\textbf{Odometry.}
The real environment may be cluttered. For the recovery strategies to be of any use, the robot must stay within a radius around the reference position in world plane,
\begin{equation*}
   \| \bar{v}_o - \hat{\bar{v}}_o \|_2,
\end{equation*}
where $\bar{\star}$ denotes the average since the previous step.

\textbf{Reference configuration.}
The robot should track the reference when no action is required to keep balance. Besides, it must not try any action if falling is inevitable at some point. It is essential to prevent dangerous motions in any situation.
\begin{equation*}
    \| q_m - \hat{q}_m \|_2.
\end{equation*}

\textbf{Foot positions and orientations.}
The feet should be flat on the ground at a suitable distance from each other. Without promoting it specifically, the policy learns to spread the legs, which is both suboptimal and unnatural. One has to work in a symmetric, local frame, to ensure this reward is decoupled from the pelvis state, which would lead to unwanted side effects otherwise. We introduce the mean foot quaternion $q_{r\text{+}l} = (q_r + q_l) / 2$ and define the relative position and orientation of the feet in this frame,
\begin{equation*}
    \presuper{r\text{+}l}{p}_{r\text{-}l} = R_{r\text{+}l}^T (p_r - p_l), \;
    \presuper[1pt]{r\text{+}l}{R}_{r\text{-}l} = R_{r\text{+}l}^T (R_r R_l^T).
\end{equation*}
The two rewards for the foot positions and orientations are
\begin{equation*}
    \| \presuper{r\text{+}l}{p}_{r\text{-}l} - \presuper{r\text{+}l}{\hat{p}}_{r\text{-}l} \|_2, \;
    \| \log(\presuper[1pt]{r\text{+}l}{R}_{r\text{-}l} \presuper[1pt]{r\text{+}l}{\hat{R}}_{r\text{-}l}^T) \|_2.
\end{equation*}

\subsubsection{Transient Dynamics}

Following a strong perturbation, recovery steps are executed to prevent falling in any case.

\textbf{Foot placement.}
We want to move the feet as soon as the CP goes outside the support polygon. We encourage moving the pelvis toward the CP to get it back under the feet,
\begin{equation*}
    \| \presuper{b}{p}_{cp} - \presuper{b}{\hat{p}}_{cp} \|_2,
\end{equation*}
where $\presuper{b}{p}_{cp}$ is the relative CP position in odometry frame.

\textbf{Dynamic stability.}
The ZMP should be kept inside the support polygon for dynamic stability,
\begin{equation*}
    \| p_{zmp} - p_{SP} \|_2,
\end{equation*}
where $SP$ is the center of the vertical projection of the footprints on the ground. It anticipates future impact with the ground and is agnostic to the contact states.

\subsubsection{Safety and Comfort}

Safety is critical during the operation of a bipedal robot. Comfort is also important for a medical exoskeleton to enhance rehabilitation.

\textbf{Balanced contact forces.}
Distributing the weight evenly on both feet is key in natural standing,
\begin{equation*}
    | F^z_r \hat{\delta}_r + F^z_l \hat{\delta}_l - m g |,
\end{equation*}
where $\hat{\delta}_r,\hat{\delta}_l \in \{0,1\}$ are the right and left contact states and $F^z_r,F^z_l$ are the vertical contact forces.

\textbf{Ground friction.}
Reducing the tangential contact forces limits internal constraints in the mechanical structure that could lead to overheating and slipping. Moreover, exploiting too much friction may lead to unrealistic behaviors,
\begin{equation*}
    \| F^{x,y}_{r,l} \|_2,
\end{equation*}
where $F^{x,y}_{r,l}$ gathers the tangential forces on the feet.

\textbf{Pelvis momentum.}
Large pelvis motion is unpleasant. Besides, reducing the angular momentum helps to keep~balance,
\begin{equation*}
    \| \bar{\omega}_b - \hat{\bar{\omega}}_b \|_2.
\end{equation*}

\subsection{Policy Conditioning for Smoothness and Safety}

Safe and predictable behavior is critical for autonomous systems evolving in human environments such as bipedal robots. Beyond this, jerky commands are not properly simulated and hardly transfer to reality, let alone it is likely to damage the hardware or shorten its lifetime. Ensuring smoothness of the policy during the training mitigates these issues without performance side effects. In the case of a medical exoskeleton, smoothness is even more critical. Vibrations can cause anxiety, and more importantly, lead to injuries over time since patients have fragile bones.

In RL, smoothness is commonly promoted through reward components, e.g.\ the total power consumption or the norm of the motor velocities \cite{Ferigo21HumanoidPush, ValkryStandingPush20}. This approach is unreliable because it is actually the return that is maximized and not the instantaneous reward. The agent would disregard any reward component that increases the likelihood of falls. Adding up regularization terms to the loss function directly gives us control over how much it is enforced during the learning.

We use temporal and spatial regularization to promote smoothness of the learned state-to-action mappings of the neural network controller. The temporal regularization term
\begin{equation*}
    L^T(\theta) =  \left\| \mu_{\theta}(s_t) - \mu_{{\theta}}(s_{t+1}) \right\|_1,
\end{equation*} with weight $\lambda_T$ as well as the spatial regularization term
\begin{equation*}
    L^S(\theta) = \| \mu_{\theta}(s_t) - \mu_{\theta}(\tilde{s}_{t}) \|_2^2,
\end{equation*}
 where $\tilde{s}_t \sim \mathcal{N}(s_t, \sigma_S)$ with weight $\lambda_S$ are added to the original surrogate loss function of the PPO algorithm $L^{ppo}$,
\begin{equation*}
    L(\theta) = L^{ppo}(\theta) + \lambda_T L^T(\theta) + \lambda_S L^S(\theta).
\end{equation*}

It was suggested in \cite{Mysore21CAPS} to choose $\sigma_S$ based on expected measurement noise and/or tolerance. However, it limits its effectiveness to robustness concerns. Its true power is unveiled when smoothness is further used to enforce regularity and cluster the behavior of the policy \cite{Shen20DRL}. By choosing the standard deviation properly, in addition to robustness, we were able to learn a minimal yet efficient set of recovery strategies, as well as to adjust the responsiveness and reactivity of the policy on the real device. A further improvement is the introduction of the L1-norm in the temporal regularization. It still guarantees that the policy reacts only if needed and recovers as fast as possible with minimal action, but it also prevents penalizing too strictly short peaks corresponding to very dynamic motions. It is beneficial to withstand strong pushes and reduce the number of steps. Finally, it is more appropriate to penalize the mean of the policy $\mu_{\theta}(s_t)$ instead of the actions $\pi_{\theta}(s_t)$ as originally stated. It provides the same gradient wrt the parameters $\theta$ but is independent of the standard deviation $\sigma$, which avoids penalizing exploration.

\begin{figure}
    \centering
    \includegraphics[trim=0.8em 1.4em 0.7em -0.5em, width=\columnwidth]{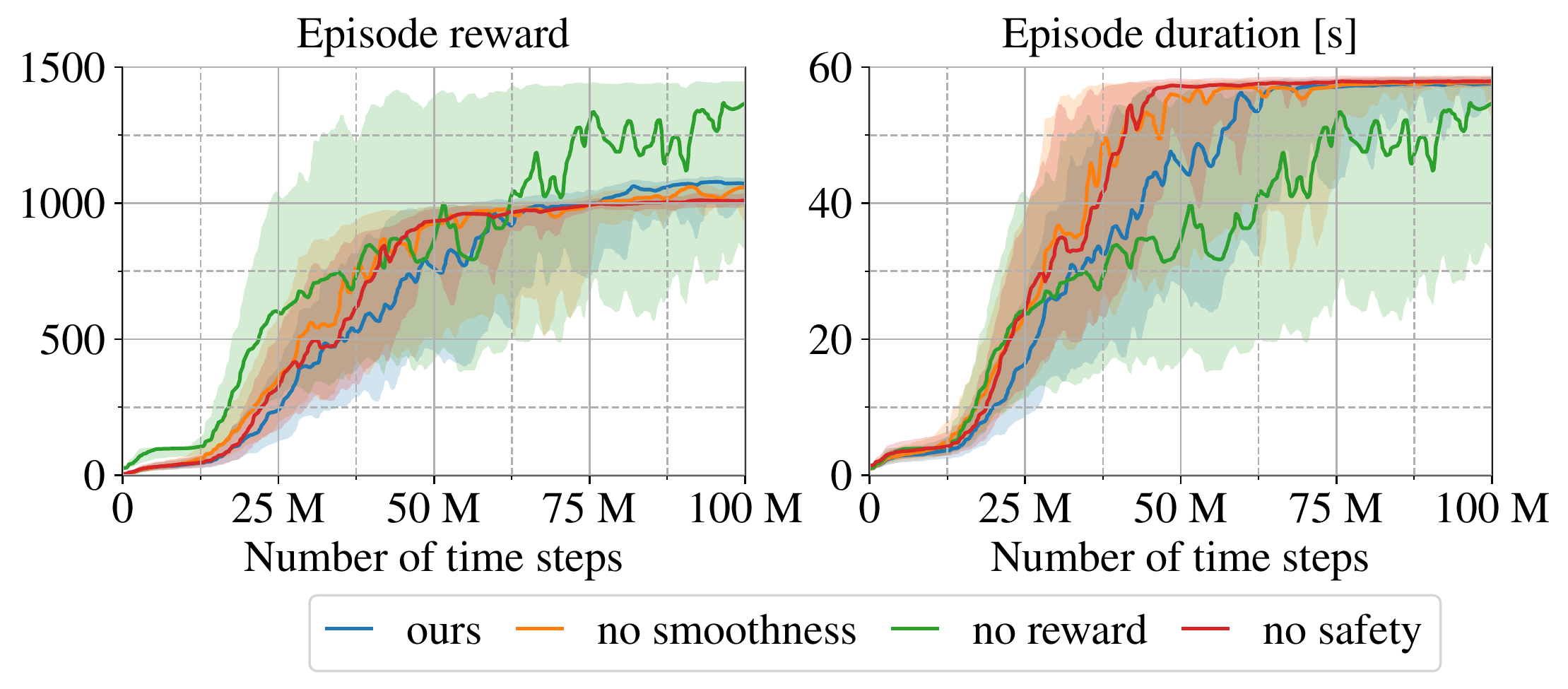}
    \caption{Ablation study of our proposed reward formulation, policy regularization for smoothness and termination conditions for safety. The standard deviation is taken over episodes per batch.}
    \vspace{-1.2em}
    \label{fig:training-analysis}
\end{figure}

\subsection{Actor and Critic Network Architecture}

The policy and the value function have the same architecture but do not share parameters. The easiest way to ensure safety is to limit their expressiveness, i.e. minimizing the number of parameters. It enhances the generalization ability by mitigating overfitting, and thereby leads to more predictable and robust behavior on the real robot, even for unseen or noisy observations. However, it hinders the overall performance of the policy. Different networks with varying depth and width have been assessed by grid search. The final architecture has 2 hidden layers with 64 units each. Below this size, the performance drops rapidly.

\section{Results}

\begin{figure}
    \centering
    \includegraphics[trim=0 0 0 -2.2em, width=\columnwidth]{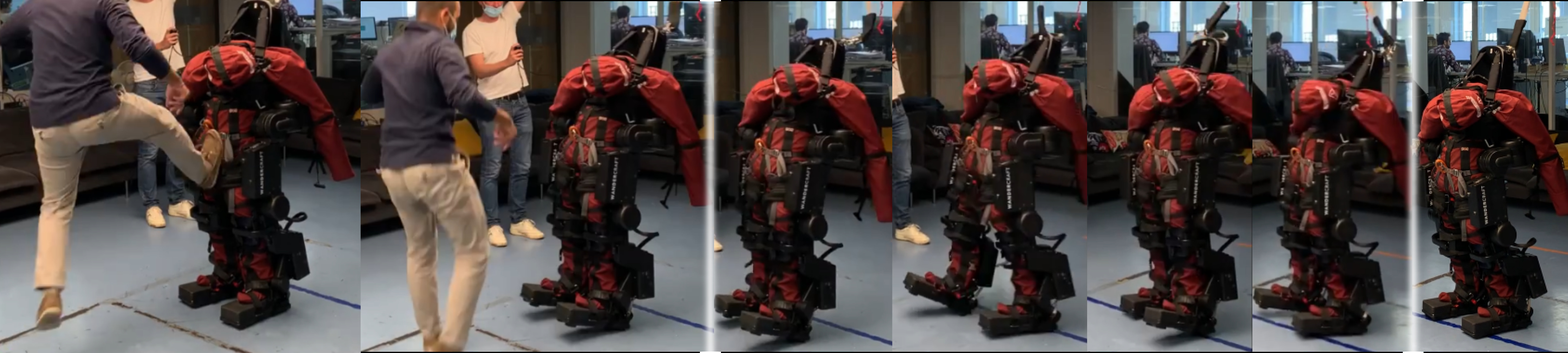}
    \caption{Strong impact kick, 1 recovery step per frame in highlighted section.}
    \vspace{-1.2em}
    \label{fig:high_kick}
\end{figure}

\subsection{Training Setup and Performance}

We train our policy using the open-source framework RLlib\cite{rllib17} with refined PPO parameters. The episode duration is limited to $60s$, which corresponds to $T=1500$ time steps. In practice, 100M steps are necessary for asymptotic optimality under worst-case conditions, corresponding to roughly four months of experience on a real robot. This takes 6h to obtain a satisfying and transferable policy, using $40$ independent simulation workers on a single machine with 64 physical cores and 1 GPU Tesla V100.

The training  curves  of  the  average episode reward and duration in \autoref{fig:training-analysis} show the impact of our main contributions:
\begin{itemize}
    \item Smoothness conditioning slightly slows down the convergence but does not impede the asymptotic reward.
    \item Without reward engineering, i.e. systematically $+1$ per step, similar training performance can be observed until 25M steps thanks to the well-defined termination conditions. After this point, the convergence gets slower, and the policy slightly underperforms at the end. This result validates our convergence robustness and that our reward provides insight into how to recover balance.
    \item Without the termination conditions for safety and transferability, faster convergence in around 30M steps is achieved. It is consistent with \cite{Li21PushCassie, Ferigo21HumanoidPush, Melo20Push}. However, using such a policy on the real device would be dangerous.
\end{itemize}

\subsection{Closed-loop Dynamics}

\autoref{fig:smoothness_conditioning} shows that smoothness conditioning improves the learned behavior, cancels harmful vibrations and preserves dynamic motions. Moreover, it also recovers balance more efficiently, by taking shorter and minimal actions.

\subsubsection{Vibrations in Standing}

Our regularization is a key element in avoiding vibrations in standing on the real device and promoting smooth actions. The effect is clearly visible in the target velocity predicted from our policy network, see \autoref{fig:smoothness_conditioning}. The target velocity without regularization leads to noisy positions and bang-bang torque commands, whereas our proposed framework learns an inherently smooth policy. Moreover, using L1 norm for the temporal regularization enables us to preserve the peaks at $13.8s$ and $17.5s$, which is critical to handle pushes and execute agile recovery motions.

\subsubsection{Steady-state Attractiveness and Hysteresis}

\begin{figure}
    \centering
    \includegraphics[trim=0.8em 1.1em 0.6em -0.8em, width=\columnwidth]{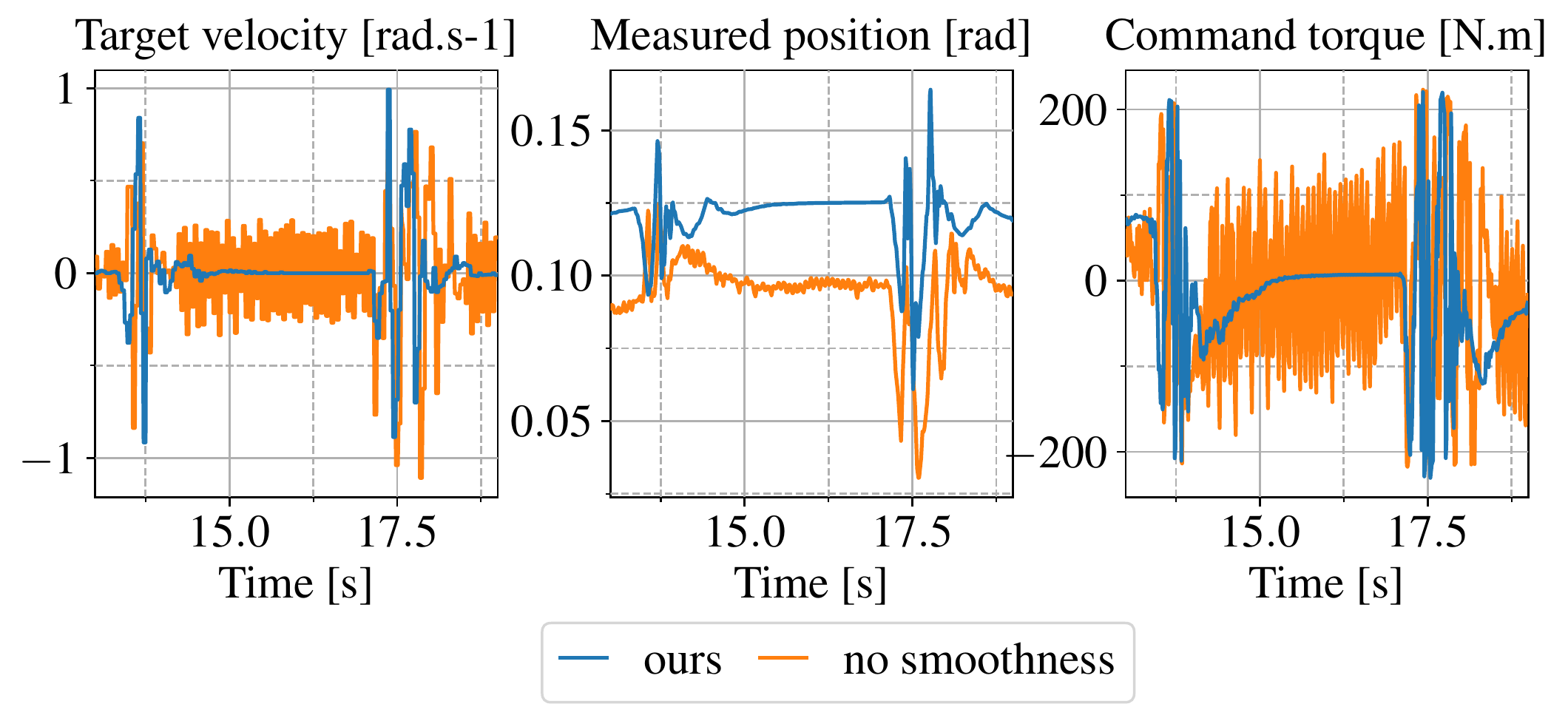}
    \caption{Comparison of predicted target velocity, measured position and command torque from the PD controller of the left knee joint with and without smoothness conditioning.}
    \vspace{-1.2em}
    \label{fig:smoothness_conditioning}
\end{figure}

Once pushed, the robot recovers balance with minimal action, see \autoref{fig:smoothness_conditioning}. It quickly goes back to standing upright close to the reference. A small hysteresis is observed. It does not affect the stability and avoids doing an extra step to move the feet back in place, see \autoref{fig:mini1}. It will vanish after the next push.

\subsection{Analysis of Learned Behaviors}

\begin{figure}
    \centering
    \includegraphics[trim=0 1.0em 0 -0.8em, width=0.59\columnwidth]{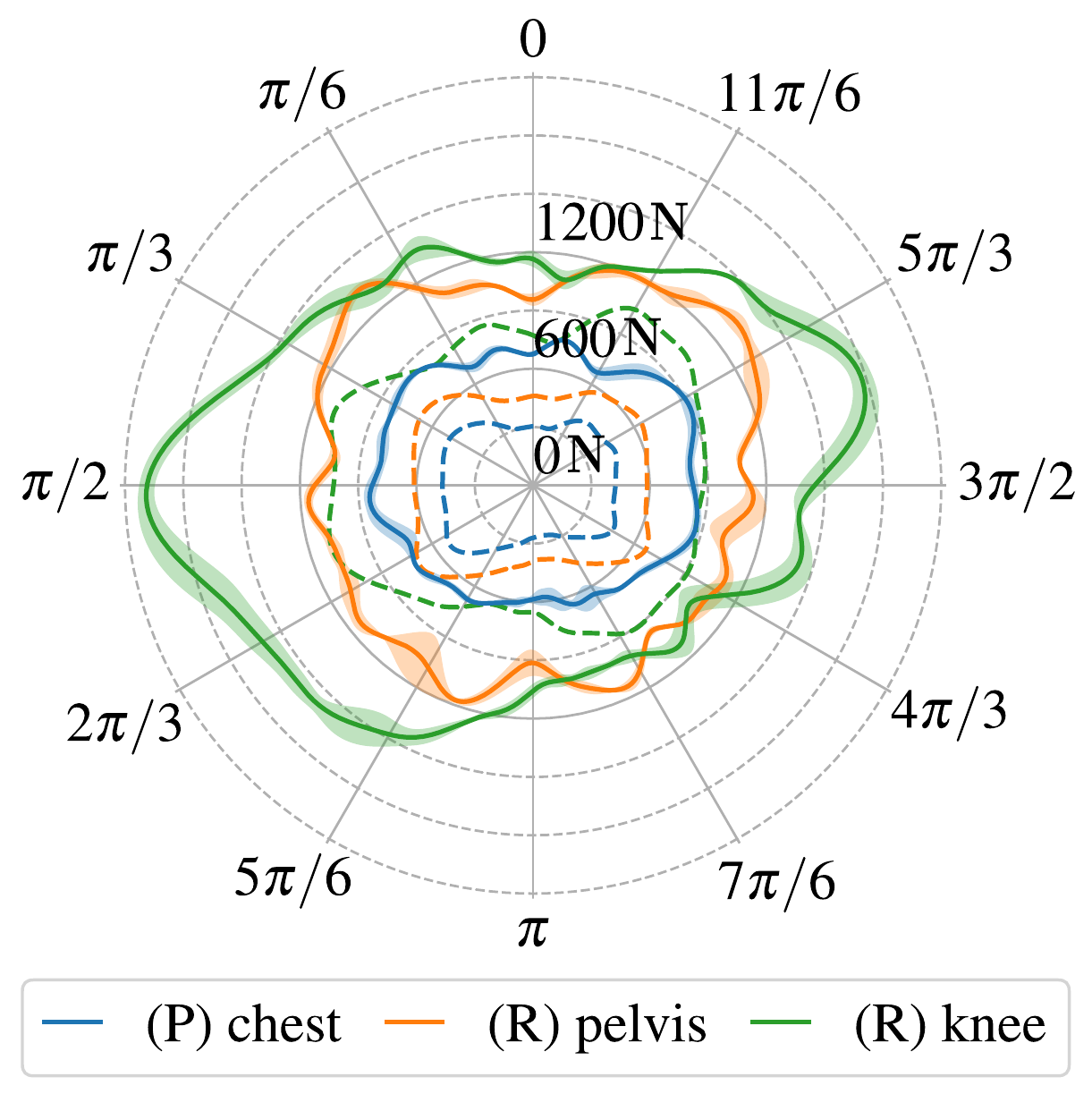}
    \caption{Maximum recoverable force magnitude $F$, applied from any direction in plane $(\cos(\phi), \sin(\phi), 0)$ at several locations on the patient (P) and the left side of the robot (R). Solid and dashed lines are respectively associated with our policy and PID tracking of the reference standing pose.}
    \vspace{-1.2em}
    \label{fig:force-analysis}
\end{figure}

Different recovery strategies are generated and tested in simulation for horizontal pushes on the pelvis. We build an interactive playground in Jiminy, to test the trained policy with pushes from all sides, see \autoref{fig:jiminy}. Depending on the direction and application point, a specific recovery is triggered and a different magnitude of force can be handled, see \autoref{fig:force-analysis}.
In-place strategies, like ankle or hip strategy, are sufficient for small pushes from any direction. Strong front, back and diagonal pushes are handled with reactive stepping strategies, and even jumps, while side pushes activate a different behavior performed with the ankles, see \autoref{tab:emerging-strategy}. For side pushes, the robot twists the stance foot alternating around the heel and the toes while balancing the opposite leg, \autoref{fig:mini5}. This dancing-like behavior avoids weight transfer, which was found the most difficult to learn.
A larger variety of push recovery strategies on the real device are displayed in the supplementary video\footref{fn:video}.

\begin{table}[!htbp]
\vspace{-0.4em}
\caption{Overview of emerging strategies for pushes from all sides}
\label{tab:emerging-strategy}
\centering
\begin{tabular}{c | c c c c c}
Emerging strategy & front & back & diagonal & side & small\\
 \hline
 \rowcolor{gray!20}
 Ankle control & \xmark & \xmark & \xmark & \cmark & \cmark \\
 Hip control & \xmark & \xmark & \xmark & \cmark & \cmark \\
 \rowcolor{gray!20}
 Stepping & \cmark & \cmark & \cmark & \xmark & \xmark \\
 Jumping & \cmark & \xmark & \cmark & \xmark & \xmark \\
 \rowcolor{gray!20}
 Foot tilting & \cmark & \cmark & \xmark & \xmark & \cmark \\
 Angular momentum & \xmark & \xmark & \xmark & \cmark & \xmark \\
\end{tabular}
\vspace{-0.5em}
\end{table}

Recent results in simulation on the Valkyrie robot \cite{ValkryStandingPush20} invite for comparison. The humanoid has roughly the same weight as the exoskeleton Atalante carrying a dummy ($135kg$) and height of an average human person ($1.8m$). To compare impulses, the applied forces are put in relation to the duration of the push. We can handle impulses of about $190Ns$ for sagittal pushes on the pelvis with our safe policy shown in \autoref{fig:force-analysis}, compared to $240Ns$ for Yang \textit{et al.} \cite{ValkryStandingPush20}. This is satisfactory considering that the motor torque limits are about 50\% lower on Atalante and knowing that the safety constraints are limiting the performance of our policy.

\subsection{Experimental Validation on the Exoskeleton Atalante}

The trained control policy is evaluated qualitatively for both valid users and dummies of different masses on several Atalante units in the Wandercraft offices. Even though the robot is only pushed at the pelvis center during the learning, \autoref{fig:force-analysis} strongly suggests that the policy can handle many types of external disturbances. We push the robot in reality at multiple application points and obtain impressive results, see \autoref{fig:mini_scenarios} and \ref{fig:high_kick}. The recovery strategies are reliable for different push variations and pulling. The transfer to Atalante works out-of-the-box despite wear of the hardware and discrepancy to simulation, notably ground friction, mechanical flexibility and patient disturbances.
\section{Conclusion and Future Works}

We obtain a controller that provides robust and versatile recovery strategies. Several techniques are combined to promote smooth, safe and predictable behaviors, even for unexpected events and unrecoverable scenarios. As theoretical guarantees are limited, our method was only verified empirically in simulation and reality. The policy was easily transferred to a medical exoskeleton. Even though trained for a single average patient model, our policy was validated experimentally with different users and dummies. It performed successfully a variety of recovery motions against unknown perturbations at various locations. To our knowledge, we are the first to demonstrate reactive push recovery in standstill on a real humanoid robot using deep RL.

Our framework being generic, theoretically it could be applied to any reference motion to stabilize it and provide recovery capability. For now, walking was put aside because it is more challenging than standing. Indeed, the stability is precarious during single support phases and the mechanical deformation of the structure becomes problematic. We are planning to unify walking and push recovery in future works. Besides, our framework can be adapted to other bipedal robots, and it would be interesting to compare the performance on other platforms. Further research directions include switching to more sample-efficient off-policy algorithms and enhancing exploration via curiosity-based intrinsic reward.

\section{Acknowledgment}

Experiments presented in this paper were carried out using the Grid'5000 testbed, supported by a scientific interest group hosted by Inria and including CNRS, RENATER and several Universities as well as other organizations (see https://www.grid5000.fr).


\bibliographystyle{IEEEtran}
\bibliography{IEEEabrv,main}

\begin{thebibliography}{10}
\providecommand{\url}[1]{#1}
\csname url@samestyle\endcsname
\providecommand{\newblock}{\relax}
\providecommand{\bibinfo}[2]{#2}
\providecommand{\BIBentrySTDinterwordspacing}{\spaceskip=0pt\relax}
\providecommand{\BIBentryALTinterwordstretchfactor}{4}
\providecommand{\BIBentryALTinterwordspacing}{\spaceskip=\fontdimen2\font plus
\BIBentryALTinterwordstretchfactor\fontdimen3\font minus
  \fontdimen4\font\relax}
\providecommand{\BIBforeignlanguage}[2]{{%
\expandafter\ifx\csname l@#1\endcsname\relax
\typeout{** WARNING: IEEEtran.bst: No hyphenation pattern has been}%
\typeout{** loaded for the language `#1'. Using the pattern for}%
\typeout{** the default language instead.}%
\else
\language=\csname l@#1\endcsname
\fi
#2}}
\providecommand{\BIBdecl}{\relax}
\BIBdecl
\renewcommand{\BIBentryALTinterwordstretchfactor}{4}

\bibitem{Caron18StairClimb}
S.~Caron, A.~Kheddar, and O.~Tempier, ``Stair climbing stabilization of the
  hrp-4 humanoid robot using whole-body admittance control,''
  \emph{International Conference on Robotics and Automation (ICRA)}, pp.
  277--283, 2019.

\bibitem{MotionPlanning13}
S.~Dalibard, A.~El~Khoury, F.~Lamiraux, A.~Nakhaei, M.~Ta{\"i}x, and J.-P.
  Laumond, ``Dynamic walking and whole-body motion planning for humanoid
  robots: an integrated approach,'' \emph{The International Journal of Robotics
  Research (IJRR)}, vol.~32, no. 9-10, pp. 1089--1103, 2013.

\bibitem{Kuindersma2016optim}
S.~Kuindersma, R.~Deits, M.~Fallon, A.~Valenzuela, H.~Dai, F.~Permenter,
  T.~Koolen, P.~Marion, and R.~Tedrake, ``Optimization-based locomotion
  planning, estimation, and control design for the atlas humanoid robot,''
  \emph{Autonomous robots}, pp. 429--455, 2016.

\bibitem{Gurriet18Wdc}
T.~Gurriet, S.~Finet, G.~Boeris, A.~Duburcq, A.~Hereid, O.~Harib, M.~Masselin,
  J.~Grizzle, and A.~D. Ames, ``Towards restoring locomotion for paraplegics:
  Realizing dynamically stable walking on exoskeletons,'' in \emph{IEEE
  International Conference on Robotics and Automation (ICRA)}, 2018, pp.
  2804--2811.

\bibitem{Moro2018}
F.~L. Moro and L.~Sentis, ``Whole-body control of humanoid robots,'' in
  \emph{Humanoid robotics: a reference}.\hskip 1em plus 0.5em minus 0.4em\relax
  Springer, 2019, pp. 1--23.

\bibitem{Herzog14}
A.~Herzog, N.~Rotella, S.~Mason, F.~Grimminger, S.~Schaal, and L.~Righetti,
  ``Momentum control with hierarchical inverse dynamics on a torque-controlled
  humanoid,'' \emph{Autonomous Robots}, vol.~40, 2014.

\bibitem{Kim20}
D.~Kim, S.~J. Jorgensen, J.~Lee, J.~Ahn, J.~Luo, and L.~Sentis, ``Dynamic
  locomotion for passive-ankle biped robots and humanoids using whole-body
  locomotion control,'' \emph{The International Journal of Robotics Research},
  vol.~39, no.~8, pp. 936--956, 2020.

\bibitem{sutton2018reinforcement}
R.~S. Sutton and A.~G. Barto, \emph{Reinforcement learning: An
  introduction}.\hskip 1em plus 0.5em minus 0.4em\relax MIT press, 2018.

\bibitem{Mysore21CAPS}
S.~Mysore, B.~Mabsout, R.~Mancuso, and K.~Saenko, ``Regularizing action
  policies for smooth control with reinforcement learning,'' \emph{IEEE
  International Conference on Robotics and Automation (ICRA)}, pp. 1810--1816,
  2021.

\bibitem{Li21PushCassie}
Z.~Li, X.~Cheng, X.~B. Peng, P.~Abbeel, S.~Levine, G.~Berseth, and K.~Sreenath,
  ``Reinforcement learning for robust parameterized locomotion control of
  bipedal robots,'' \emph{IEEE International Conference on Robotics and
  Automation (ICRA)}, pp. 2811--2817, 2021.

\bibitem{Ferigo21HumanoidPush}
D.~Ferigo, R.~Camoriano, P.~M. Viceconte, D.~Calandriello, S.~Traversaro,
  L.~Rosasco, and D.~Pucci, ``On the emergence of whole-body strategies from
  humanoid robot push-recovery learning,'' \emph{IEEE/RAS International
  Conference on Humanoid Robots (Humanoids)}, 2020.

\bibitem{Melo20Push}
D.~C. Melo, M.~R. O.~A. Máximo, and A.~M. da~Cunha, ``Push recovery strategies
  through deep reinforcement learning,'' in \emph{2020 Latin American Robotics
  Symposium (LARS)}, 2020, pp. 1--6.

\bibitem{Yuan20BalanceControl}
K.~Yuan, C.~McGreavy, C.~Yang, W.~Wolfslag, and Z.~Li, ``Decoding motor skills
  of artificial intelligence and human policies: A study on humanoid and human
  balance control,'' \emph{IEEE Robotics \& Automation Magazine (RA-M)},
  vol.~27, no.~2, pp. 87--101, 2020.

\bibitem{Hyon09}
S.-H. Hyon, R.~Osu, and Y.~Otaka, ``Integration of multi-level postural
  balancing on humanoid robots,'' in \emph{IEEE International Conference on
  Robotics and Automation (ICRA)}, 2009, pp. 1549--1556.

\bibitem{Stephens10}
B.~J. Stephens and C.~G. Atkeson, ``Dynamic balance force control for compliant
  humanoid robots,'' in \emph{IEEE/RSJ International Conference on Intelligent
  Robots and Systems (IROS)}, 2010, pp. 1248--1255.

\bibitem{Li17FootTilt}
Z.~Li, C.~Zhou, Q.~Zhu, and R.~Xiong, ``Humanoid balancing behavior featured by
  underactuated foot motion,'' \emph{IEEE Transactions on Robotics (T-RO)},
  vol.~33, no.~2, pp. 298--312, 2017.

\bibitem{Caron19}
S.~Caron, ``Biped stabilization by linear feedback of the variable-height
  inverted pendulum model,'' \emph{IEEE International Conference on Robotics
  and Automation (ICRA)}, pp. 9782--9788, 2020.

\bibitem{Kajita03}
S.~Kajita, F.~Kanehiro, K.~Kaneko, K.~Fujiwara, K.~Harada, K.~Yokoi, and
  H.~Hirukawa, ``Biped walking pattern generation by using preview control of
  zero-moment point,'' \emph{IEEE International Conference on Robotics and
  Automation (ICRA)}, vol.~2, pp. 1620--1626, 2003.

\bibitem{MPC06Wieber}
P.-b. Wieber, ``Trajectory free linear model predictive control for stable
  walking in the presence of strong perturbations,'' in \emph{IEEE/RAS
  International Conference on Humanoid Robots (Humanoids)}, 2006.

\bibitem{3DLip}
S.~Kajita, F.~Kanehiro, K.~Kaneko, K.~Yokoi, and H.~Hirukawa, ``The 3d linear
  inverted pendulum mode: a simple modeling for a biped walking pattern
  generation,'' in \emph{IEEE/RSJ International Conference on Intelligent
  Robots and Systems (IROS)}, vol.~1, 2001, pp. 239--246.

\bibitem{LillicrapHPHETS15}
T.~P. Lillicrap, J.~J. Hunt, A.~Pritzel, N.~Heess, T.~Erez, Y.~Tassa,
  D.~Silver, and D.~Wierstra, ``Continuous control with deep reinforcement
  learning,'' in \emph{International Conference on Learning Representations
  (ICLR)}, Y.~Bengio and Y.~LeCun, Eds., 2016.

\bibitem{HeessTSLMWTEWER17}
\BIBentryALTinterwordspacing
N.~Heess \emph{et~al.}, ``Emergence of locomotion behaviours in rich
  environments,'' 2017. [Online] \url{https://arxiv.org/abs/1707.02286}
\BIBentrySTDinterwordspacing

\bibitem{Hwangbo2019}
J.~Hwangbo, J.~Lee, A.~Dosovitskiy, D.~Bellicoso, V.~Tsounis, V.~Koltun, and
  M.~Hutter, ``Learning agile and dynamic motor skills for legged robots,''
  \emph{Science Robotics}, vol.~4, 2019.

\bibitem{Miki2022LearningRP}
T.~Miki, J.~Lee, J.~Hwangbo, L.~Wellhausen, V.~Koltun, and M.~Hutter,
  ``Learning robust perceptive locomotion for quadrupedal robots in the wild,''
  \emph{Science Robotics}, vol.~7, 2022.

\bibitem{Peng18DeepMimic}
X.~B. Peng, P.~Abbeel, S.~Levine, and M.~van~de Panne, ``Deepmimic:
  Example-guided deep reinforcement learning of physics-based character
  skills,'' \emph{ACM Transactions on Graphics (TOG)}, pp. 1--14, 2018.

\bibitem{Peng2020}
X.~Peng, E.~Coumans, T.~Zhang, T.-W. Lee, J.~Tan, and S.~Levine, ``Learning
  agile robotic locomotion skills by imitating animals,'' in \emph{Robotics:
  Science and Systems (RSS)}, 2020.

\bibitem{Rodriguez2021}
D.~Rodriguez and S.~Behnke, ``Deepwalk: Omnidirectional bipedal gait by deep
  reinforcement learning,'' in \emph{IEEE International Conference on Robotics
  and Automation (ICRA)}, 2021, pp. 3033--3039.

\bibitem{ValkryStandingPush20}
C.~Yang, K.~Yuan, W.~Merkt, T.~Komura, S.~Vijayakumar, and Z.~Li, ``Learning
  whole-body motor skills for humanoids,'' in \emph{IEEE/RAS International
  Conference on Humanoid Robotics (Humanoids)}, 2018.

\bibitem{Yang20HumanBias}
C.~Yang, K.~Yuan, S.~Heng, T.~Komura, and Z.~Li, ``Learning natural locomotion
  behaviors for humanoid robots using human bias,'' \emph{IEEE Robotics and
  Automation Letters (RA-L)}, vol.~5, pp. 2610--2617, 2020.

\bibitem{Xie2019}
Z.~Xie, P.~Clary, J.~Dao, P.~Morais, J.~Hurst, and M.~van~de Panne, ``Learning
  locomotion skills for cassie: Iterative design and sim-to-real,'' in
  \emph{Conference on Robot Learning (CoRL)}, ser. Proceedings of Machine
  Learning Research, vol. 100.\hskip 1em plus 0.5em minus 0.4em\relax PMLR,
  2020, pp. 317--329.

\bibitem{CassieStairClimb}
J.~Siekmann, K.~Green, J.~Warila, A.~Fern, and J.~W. Hurst, ``Blind bipedal
  stair traversal via sim-to-real reinforcement learning,'' in \emph{Robotics:
  Science and Systems (RSS)}, 2021.

\bibitem{Castillo2021RobustFM}
G.~A. Castillo, B.~Weng, W.~Zhang, and A.~Hereid, ``Robust feedback motion
  policy design using reinforcement learning on a 3d digit bipedal robot,''
  \emph{IEEE/RSJ International Conference on Intelligent Robots and Systems
  (IROS)}, pp. 5136--5143, 2021.

\bibitem{PPOSchulmanWDRK17}
\BIBentryALTinterwordspacing
J.~Schulman, F.~Wolski, P.~Dhariwal, A.~Radford, and O.~Klimov, ``Proximal
  policy optimization algorithms,'' 2017. [Online]
  \url{https://arxiv.org/abs/1707.06347}
\BIBentrySTDinterwordspacing

\bibitem{schulman2018highdimensional}
J.~Schulman, P.~Moritz, S.~Levine, M.~Jordan, and P.~Abbeel, ``High-dimensional
  continuous control using generalized advantage estimation,'' in
  \emph{International Conference on Learning Representations (ICLR)}, 2016.

\bibitem{Jiminy}
\BIBentryALTinterwordspacing
A.~Duburcq, ``Jiminy: a fast and portable python/c++ simulator of
  poly-articulated systems for reinforcement learning,'' 2019. [Online]
  \url{https://github.com/duburcqa/jiminy}
\BIBentrySTDinterwordspacing

\bibitem{pinocchioweb}
J.~Carpentier, F.~Valenza, N.~Mansard \emph{et~al.}, ``Pinocchio: fast forward
  and inverse dynamics for poly-articulated systems,''
  https://stack-of-tasks.github.io/pinocchio, 2015--2021.

\bibitem{MuJoCo}
E.~Todorov, T.~Erez, and Y.~Tassa, ``Mujoco: A physics engine for model-based
  control,'' in \emph{IEEE/RSJ International Conference on Intelligent Robots
  and Systems (IROS)}, 2012, pp. 5026--5033.

\bibitem{Vigne20}
M.~Vigne, A.~E. Khoury, F.~D. Meglio, and N.~Petit, ``State estimation for a
  legged robot with multiple flexibilities using imus: A kinematic approach,''
  \emph{IEEE Robotics and Automation Letters (RA-L)}, 2020.

\bibitem{ma2021spacetimeBounds}
L.-K. Ma, Z.~Yang, X.~Tong, B.~Guo, and K.~Yin, ``Learning and exploring motor
  skills with spacetime bounds,'' pp. 251--263, 2021.

\bibitem{Shen20DRL}
Q.~Shen, Y.~Li, H.~Jiang, Z.~Wang, and T.~Zhao, ``Deep reinforcement learning
  with robust and smooth policy,'' in \emph{International Conference on Machine
  Learning (ICML)}, ser. Proceedings of Machine Learning Research, vol.
  119.\hskip 1em plus 0.5em minus 0.4em\relax PMLR, 2020, pp. 8707--8718.

\bibitem{rllib17}
E.~Liang, R.~Liaw, R.~Nishihara, P.~Moritz, R.~Fox, K.~Goldberg, J.~Gonzalez,
  M.~Jordan, and I.~Stoica, ``Rllib: Abstractions for distributed reinforcement
  learning,'' in \emph{Proceedings of the 35th International Conference on
  Machine Learning}, ser. Proceedings of Machine Learning Research,
  vol.~80.\hskip 1em plus 0.5em minus 0.4em\relax PMLR, 2018, pp. 3053--3062.

\end{thebibliography}

\end{document}